\documentclass[fleqn,10pt]{wlscirep}
\usepackage[utf8]{inputenc}
\usepackage[T1]{fontenc}
\usepackage{lineno}

\usepackage{times}
\usepackage{helvet}
\usepackage{courier}

\usepackage{amsmath,amssymb,amsfonts}
\usepackage{algorithmic}
\usepackage{graphicx}
\usepackage{textcomp}
\usepackage{xcolor}
\usepackage{caption}
\usepackage{subcaption}
\graphicspath{ {./Figures/} }
\usepackage{tabularx}
\usepackage{hyperref}
\usepackage{algorithm}
\usepackage{algorithmic}
\usepackage{multirow}

\usepackage{cite}

\usepackage{makecell}

\usepackage{xcolor}
\usepackage{soul}
\usepackage{colortbl}
\usepackage{comment}

\usepackage{tikz}
\usepackage{pgfplots}
\pgfplotsset{compat=1.17}

\title{Analyzing Poverty through Intra-Annual Time-Series: A Wavelet Transform Approach}

\author[1,4,*,+]{Mohammad Kakooei}
\author[1,4,+]{Klaudia Solska}
\author[1,2,3,4]{Adel Daoud}

\affil[1]{Department of Computer Science and Engineering, Chalmers University of Technology, Gothenburg, Sweden}
\affil[2]{Institute for Analytical Sociology, Linköping University, Sweden}
\affil[3]{Center for Advanced Study in the Behavioral Sciences, Stanford University, United States}
\affil[4]{The AI and Global Development Lab (www.global-lab.ai)}

\affil[*]{corresponding author: Mohammad Kakooei (kakooei@chalmers.se)}

\affil[+]{these authors contributed equally to this work}


\begin{abstract}
Reducing global poverty is a key objective of the Sustainable Development Goals (SDGs). Achieving this requires high-frequency, granular data to capture neighborhood-level changes, particularly in data scarce regions such as low- and middle-income countries. To fill in the data gaps, recent computer vision methods combining machine learning (ML) with earth observation (EO) data to improve poverty estimation. However, while much progress have been made, they often omit intra-annual variations, which are crucial for estimating poverty in agriculturally dependent countries. We explored the impact of integrating intra-annual NDVI information with annual multi-spectral data on model accuracy. To evaluate our method, we created a simulated dataset using Landsat imagery and nighttime light data to evaluate EO-ML methods that use intra-annual EO data. Additionally, we evaluated our method against the Demographic and Health Survey (DHS) dataset across Africa. Our results indicate that integrating specific NDVI-derived features with multi-spectral data provides valuable insights for poverty analysis, emphasizing the importance of retaining intra-annual information.
\end{abstract}
\begin{document}

\flushbottom
\maketitle
%
%
\thispagestyle{empty}


\section*{Introduction}

{A}{ccurately} estimating poverty is essential for understanding socioeconomic disparities, guiding policy decisions, and tracking progress toward the United Nations Sustainable Development Goals (SDGs) \cite{SDG}. Among these goals, the eradication of poverty is a primary objective, particularly emphasized in Goal 1, No poverty \cite{SDG-poverty}. Reducing global poverty, particularly in low- and middle-income countries, is a central objective of SDGs. Achieving this goal requires policymakers to have access to reliable, high-resolution geo-temporal data that can track changes at the neighborhood level \cite{daoud2023using,daoud_what_2016,balgi2022personalized}. Poverty research is crucial because it illuminates the living conditions of marginalized populations, enabling the design of targeted and effective interventions. Despite the importance of such data, policymakers often rely on censuses and household surveys, such as the Demographic and Health Surveys (DHS) \cite{DHS_Program} and the Living Standards Measurement Study (LSMS)\cite{LSMS}, which are infrequently collected and costly to implement, leading to a scarcity of timely poverty data essential for monitoring progress toward the SDGs \cite{sustainable2015data,sandefur2015political,daoud2016association}.

Traditional poverty surveys, while valuable, are often prohibitively expensive and logistically challenging, particularly in remote or economically disadvantaged areas where continuous data collection is rare \cite{daoud2020statistical}. In response to these limitations, there has been growing interest in combining machine learning (ML) with earth observation (EO) data to estimate poverty at a neighborhood level \cite{piaggesi2019predicting,jean2016combining,ayush2020efficient,zhao2019estimation,yeh2020using,huang2021saturated,kino2021scoping,kakooei2024increasing}. Although these emerging methods offer promise, the resulting poverty maps are often insufficiently accurate for precise policy applications.

Current state-of-the-art models typically aggregate data over multiple years to address challenges such as missing data, noise from cloud cover, seasonal variations, and satellite instrument errors. Researchers often mitigate these issues by summarizing data through techniques like median value computation over extended periods. For example Yeh et.al. \cite{yeh2020using} and Pettersson et al. \cite{pettersson2023time} used three-year median of Landsat data. While this approach effectively smooths out noise and reduces data volume, it risks overlooking critical intra-annual variations. These temporal variations—such as fluctuations in agricultural activity \cite{tang2022predicting,tang2018dynamic} and economic events \cite{ratledge2022using}—can provide valuable insights that enhance the accuracy of poverty predictions if properly accounted for. Thus, a balance must be struck between simplifying data for model manageability and preserving temporal detail crucial for accurate poverty estimation.

The aim of this article is to address this question with the goal of improving poverty estimation accuracy that incorporates intra-annual time-series data. The article demonstrates the utility of wavelet-based feature extraction using a simulated dataset, where nighttime light data serves as a proxy for poverty estimation. Nighttime light, which correlates with economic activity, offers a unique and accessible measure for assessing poverty. The simulation uses nighttime light data as the target variable, representing a wealth index, while input data is derived from Landsat multispectral satellite images. Following this, the study applies the proposed methodology to real-world data, using the DHS dataset as a case study across the African continent. In this context, the time-series Normalized Difference Vegetation Index (NDVI), alongside raw Landsat spectral bands, is used in training a deep learning model to enhance the precision of poverty estimates. NDVI has demonstrated strong potential as an independent variable in economic analyses, particularly in agriculture-dependent economies \cite{tang2022predicting,JOHNSON2014}.

The other aspect of this research lies in its use of the wavelet transform as a tool for temporal data summarization. Unlike traditional methods that may overlook important intra-annual variations, the wavelet transform captures and summarizes temporal information at multiple scales, retaining critical details without the need to process the entire temporal dataset. By leveraging these temporal dynamics, this study aims to refine poverty estimation models and improve their predictive performance. This approach provides a novel framework for summarizing and analyzing temporal satellite data, potentially leading to more accurate and granular assessments of poverty and other socioeconomic indicators. Moreover, the integration of ML techniques with satellite data allows for the analysis of large datasets to identify economic indicators like nighttime lights, agricultural productivity, and infrastructure development. These indicators can generate high-resolution, timely poverty estimates, providing policymakers with the detailed data necessary to address socioeconomic challenges effectively.

\textbf{Poverty estimation. } Datasets used in poverty studies encompass a wide range of sources, including household surveys, census data, administrative records, and satellite imagery. Satellite datasets are essential for poverty estimation, offering critical spatial and temporal information. Planetary-scale satellite imagery, as opposed to surveys, is accessible throughout Africa across a broad time span and geographic area. For instance, nighttime lights data from satellites like VIIRS show urbanization and economic activity levels \cite{henderson-2012} \cite{gosh-2013} \cite{Henderson2017} \cite{Bruederle-2018}. Bruederle et al. \cite{Bruederle-2018} developed a simple linear regression model combining nighttime lights data with demographic surveys, controlling for factors like population density and electrification.

EO-ML methods have been on the rise the last decades \cite{burke2021using}, but their use for poverty estimation is recent. Elvidge et al. \cite{elvidge2009global} were among the first researchers that deployed EO data to generate a poverty map. They defined poverty index as the value of population count divided by DMSP-OLS nighttime light value. Then, the index was calibrated by the national level poverty data from the World Development Indicators (WDI) 2006 edition. Wang et al. \cite{wang2012poverty} used Principal Component Analysis (PCA) to extract the poverty index IPI from 17 socio-economic indexes. They showed that there is a high correlation between IPI and DMSP-OLS nighttime light data at a province-scale in China. Furthermore, Li et al. \cite{li2013detecting} showed that they can identify a high correlation rate between DMSP NL and Gross Domestic Product (GDP) in Zimbabwe for the period 1992 to 2009 when there was an economic decline.

Jean et.al.\cite{jean2016combining} deployed 3-band high resolution optical (daylight satellite) data as input, for estimating poverty. They proposed using a deep-learning method for poverty estimation, and trained a ResNet18 deep model \cite{he2016deep} with DMSP/OLS (NL) as the output. Then the deep network was frozen and a ridge regression layer was added to the network to predict the poverty index. The model was evaluated on five African countries including, Nigeria, Tanzania, Uganda, Malawi, and Rwanda. But to evaluate the proposed method between countries, the asset index was normalized for each country as a pre-processing step. Therefore, they ignore the wealth difference between countries and the temporal variability, and the model can show just the local variation of wealth index within each country. In a similar work Ni et al. \cite{ni2020investigation} trained daylight images from Google Static Map API on NL as target data. They used VGG-Net, Inception-Net, ResNet, and DenseNet, to extract features from optical imagery and then applied LASSO regression for poverty prediction. They evaluated their method on four African countries including Malawi, Rwanda, Uganda, and Nigeria. In addition, Piaggesi et al. \cite{piaggesi2019predicting} followed a similar approach by deploying ResNet50 and VGG-F as deep models that were feed to a ridge regression model. They evaluated their method on a local-scale in Santiago (Chile), Los Angeles, Philadelphia, Boston, Chicago, and Houston (US). Building on Jean et al.'s \cite{jean2016combining} method,  Perez et al. \cite{perez2017poverty}, but included additional experiments. For example they used multi-spectral Landsat-7 data instead of the optical data. More over, they used ResNet-34 and VGG-F deep models besides the ResNet-18 model.

Tange et al. \cite{tang2022predicting} used moderate-resolution vegetation index NDVI as an evidence of wealth level in low-income countries which are heavily dependent on agriculture. They also evaluated their method on Malawi, Nigeria, Rwanda, Tanzania, and Uganda. Yeh et.al. \cite{yeh2020using} deployed to ResNet18 networks, in which one was trained using Landsat optical data and the other was trained using NL data. Their outputs were concatenated and passed through a ridge regression layer to predict the poverty in several African countries. Furthermore, Chi et.al. \cite{chi2022microestimates}  fused satellite data, mobile phone networks, topographic maps, and facebook data to predict Relative Wealth Index (RWI) in several low- and middle-income countries. They generated the poverty map at 2.4 km resolution. 
Vegetation indices from Moderate Resolution Imaging Spectroradiometer (MODIS), such as Normalized Difference Vegetation Index (NDVI) and Enhanced Vegetation Index (EVI), indicate agricultural productivity \cite{JOHNSON2014}, which can be used for better poverty prediction. Although there is now a rich literature on using EO-ML methods for poverty estimation, none of the studies previously mentioned how much intra-annual data, such monthly variation, is informative for prediction accuracy. 


\textbf{The importance of vegetation for poverty estimation. }
Time-series feature extraction from Earth observation data is crucial for understanding and predicting various phenomena, including poverty dynamics, agricultural productivity, and environmental changes. Time-series data, such as vegetation indexes, provide valuable insights into changes in environmental conditions over time, which are crucial for understanding socio-economic dynamics, especially in developing regions heavily reliant on agriculture. Tange et al. \cite{tang2022predicting} demonstrated the effectiveness of utilizing the NDVI, derived from moderate-resolution satellite imagery (MODIS), in predicting poverty indicators among agricultural communities.

The NDVI, a widely-used vegetation index, offers a measure of vegetation greenness and health, which is particularly relevant for regions where agriculture plays a significant role in livelihoods. By leveraging convolutional neural networks (CNNs) and transfer learning techniques, Tange et al. demonstrate how NDVI time-series data can be effectively harnessed to predict poverty measures such as consumption expenditure and wealth index at the community level. Tange et al.'s \cite{tang2022predicting} approach involves fine-tuning a pre-trained CNN model on NDVI images to predict nighttime light intensities, which serve as intermediate labels. Subsequently, random forest regression models are trained on the extracted NDVI features to predict poverty indicators. This two-step procedure capitalizes on the temporal dynamics captured by NDVI time-series data and the spatial information provided by nighttime light intensities, resulting in accurate and timely poverty predictions. Moreover, the authors highlight the importance of considering the temporal dimension in poverty prediction by demonstrating the ability of their model to capture changes in consumption expenditure over time among poor communities. This sequential prediction aspect adds a valuable dimension to early warning systems and policy evaluation efforts.

In summary, time-series feature extraction from EO data, particularly using NDVI, is crucial for understanding and predicting socio-economic and environmental phenomena in agricultural regions. Previous studies have highlighted the importance of capturing temporal NDVI dynamics for poverty prediction, providing valuable insights for policy evaluation. In the next subsection, we will further explore the role of NDVI in this context.

\textbf{Time-series feature extraction in EO data. } The importance of time-series EO data analysis has been demonstrated in a range of applications, including wetland mapping \cite{mohseni2023wetland,amani2022forty}, change detection \cite{amani2021wetland}, cropland analysis and yield estimation \cite{xu2021cotton}, semantic segmentation \cite{paolini2022classification}, and vegetation phenological studies \cite{zeng2020review}, which encompass both intra-annual and inter-annual variation studies. Among these applications, cropland analysis, traditionally reliant on NDVI, is particularly significant due to its strong correlation with agricultural economic analysis.

Wavelet transform is one of the key methods used for decomposing time-series data. Unlike Empirical Mode Decomposition (EMD), which does not rely on predefined basis functions, wavelet transform uses a set of predefined basis functions to achieve data decomposition \cite{ben2018comparative}. Karthikeyan et al. \cite{karthikeyan2013predictability} conducted a comparison between wavelet-based methods and EMD for time-series modeling and forecasting. Their findings showed that wavelet-based methods outperformed EMD, demonstrating superior predictive accuracy, particularly in forecasting rainfall 12 months in advance across multiple locations.

Martinez et al. \cite{martinez2009vegetation} used wavelet transform to perform non-stationary and multiscale analysis of NDVI time series, capturing both short- and long-term vegetation variations. Their study emphasized how multi-resolution analysis enables the differentiation between intra-annual and inter-annual changes, identifying critical phenological features such as minimum NDVI values and the timing of peak vegetation. Similarly, Yan et al. \cite{yan2022comparison} compared intra-annual NDVI with Annual Maximum NDVI (NDVImax) and found that time-series NDVI provided a more detailed representation of vegetation changes, revealing greater spatial and temporal heterogeneity and stronger correlations with climatic factors like precipitation and temperature.

Rhif et al. \cite{rhif2021improved} focused on improving trend analysis of non-stationary NDVI time series, particularly in monitoring long-term vegetation changes. They used Multi-Resolution Analysis Wavelet Transform (MRA-WT) to decompose NDVI series, and proposed a combined mother wavelet approach to better track vegetation trends, finding evidence of forest degradation and cropland improvement. In their subsequent work, Rhif et al. \cite{Rhif2021} introduced a hybrid approach, I-WT-LSTM (Improved Wavelet Long Short-Term Memory), which outperformed traditional models in forecasting non-stationary time series. Another study by Rhif et al. \cite{rhif2022optimal} optimized wavelet transform parameters for analyzing NDVI time series in the Mediterranean, determining that a decomposition level of 5 provided the best results in capturing trends and seasonal variations.

These studies highlight the effectiveness of wavelet-based methods for feature extraction from time-series satellite data. Intra-annual NDVI analysis enables researchers to examine how various vegetation types respond to seasonal changes, which is essential for understanding vegetation phenology. Building on this, we applied NDVI time-series data in our study to extract more detailed and informative features for economic analysis. By utilizing wavelet transformation, we aimed to enhance the analysis of NDVI time-series, providing a richer dataset for economic studies.

\section*{Results}

Echoing the focus on Africa by previous researchers \cite{jean2016combining,yeh2020using,chi2022microestimates}, our study also concentrates on African countries. As previously discussed, the primary objective of this study is to generate a more accurate poverty map of the African continent using the DHS survey dataset. However, because the DHS dataset undergoes privacy protection methods before being released, it is beneficial to create an additional dataset for simulation purposes. To do so, we employ two main data sources for our experimental investigation: a "simulated dataset" and the "DHS dataset." Before we outline our experimental setup, we review the data collections, including satellite imagery and survey points, used in this investigation. We then describe how these data are integrated to form the basis of each experimental investigation.

\subsection*{Earth Observation Datasets}

\begin{figure*}
    \centering
    \includegraphics[width=0.8\linewidth]{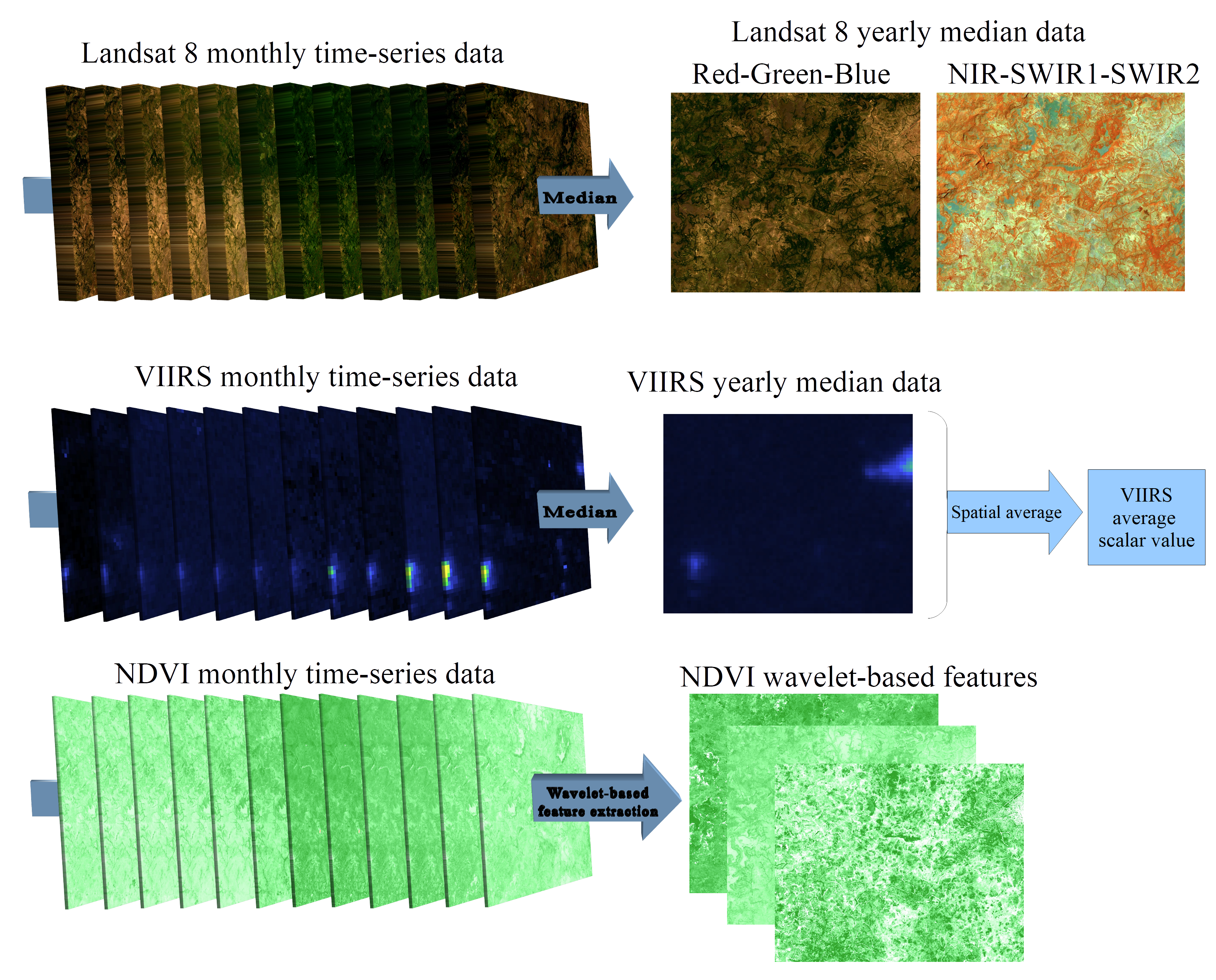}
    \caption{Data collections and satellite imagery used in this study: (a) Landsat-8 monthly time-series data, featuring 7 bands: red, green, blue, NIR, SWIR1, SWIR2, and Thermal. (b) Aggregated Landsat-8 yearly data, generated using a pixel-based median operator; the left images display RGB visualization, while the right images show false color visualization using NIR-SWIR1-SWIR2. (c) VIIRS monthly composite of nighttime light data, consisting of a single band showing the calibrated monthly average value. (d) VIIRS yearly median value image. (e) VIIRS image spatial average value. (f) NDVI monthly time-series data, each image containing one band showing the NDVI average for that month. (g) Feature extraction from NDVI time-series data using the wavelet method.}
    \label{fig:DataCollections}
\end{figure*}

Figure \ref{fig:DataCollections} shows an overview summary of the datasets that are used in this study, which includes Landsat-8 data, Visible Infrared Imaging Radiometer Suite (VIIRS) nighttime light data, and NDVI time-series datasets. Landsat 8 \cite{USGS_Landsat} is a satellite mission operated by NASA and the US Geological Survey (USGS) as part of the Landsat program. Launched in February 2013, Landsat 8 carries the Operational Land Imager (OLI) and the Thermal InfraRed Sensor (TIRS). These instruments capture multispectral imagery across visible, near-infrared, and thermal wavelengths, providing valuable data for applications such as land cover monitoring, environmental assessment, and natural resource management. The VIIRS \cite{VIIRS} is a nighttime light satellite imagery program that is available from 2014 and provides data in 500m spatial resolution. We accessed to the Landsat-8 surface reluctance and the VIIRS NL data through Google Earth Engine platform \cite{Google_Earth_Engine}. The NDVI serves as a widely-utilized metric for assessing vegetation health and density. It is calculated from Landsat-8 satellite imagery by computing the ratio between the near-infrared (NIR) and red spectral bands as follows:
\begin{equation}
    NDVI = \frac{NIR-Red}{NIR+Red}
\end{equation}

\subsection*{The Demographic and Health Surveys}
The Demographic and Health Surveys (DHS) program has been collecting data on living conditions in numerous countries, including several across Africa. In this study, the International Wealth Index (IWI), derived from DHS survey data, is used as the target variable for training deep learning models \cite{smits2015international}. The DHS program provides publicly available survey data, a crucial resource for analyzing health and living conditions in low- and middle-income countries. The DHS dataset has been widely used for studying socioeconomic factors worldwide.

To protect the privacy of surveyed households, DHS implements two levels of data pre-processing. First, it aggregates samples from the same enumeration area (EA) into a single GPS location, representing the EA's center, providing a neighborhood-level poverty estimate. Second, the EA's center is randomly displaced to further protect privacy, with a maximum displacement of 2 km in urban areas and 5 km in rural areas \cite{burgert2013geographic}. 

The independent variables in this study are a combination of VIIRS nighttime light data and Landsat 8 data, which are concatenated and used as input to the deep learning model. Given the limited availability of DHS survey points, these data are valuable and should not be excluded due to the unavailability of independent variables like Landsat 8 data. To address potential data gaps, we use the median value of Landsat 8 and nighttime light data over three consecutive years. This approach allows us to preserve the temporal variation of NDVI values by compressing the monthly data over three years into a single yearly structure. This method ensures that we capture relevant temporal dynamics while accommodating limitations in data availability.

\subsection*{Simulated dataset}
Using the mean radiance value from VIIRS nighttime light data as a proxy for poverty estimation in this simulated setting allows us to explore and evaluate our methodology for improved poverty prediction under controlled conditions, while avoiding the logistical complexities of utilizing DHS \cite{DHS_Program} household survey data. Although household survey data, such as DHS, are typically employed for poverty estimation due to their comprehensive socioeconomic indicators, privacy concerns and data access limitations associated with DHS present challenges for our research. As an alternative, the mean pixel value from nighttime light data offers a viable substitute in this simulated scenario. Nighttime light data are readily available and provide spatially and temporally comprehensive coverage globally. Moreover, the correlation between artificial nighttime light intensity and socioeconomic factors, including poverty levels, has been extensively documented in the literature \cite{henderson-2012,gosh-2013,Henderson2017,Bruederle-2018}. Additionally, other studies, such as those by Jean et al. \cite{jean2016combining}, have used nighttime light data as an intermediate target variable in deep learning models, where features extracted from the data are later used in a linear regression to predict the wealth index. These experiments confirm that nighttime light data is a valuable and informative proxy for economic indicators, supporting its use in poverty estimation and other socioeconomic analyses.

The independent variable in the simulated dataset is Landsat 8 data, which is used to estimate the average nighttime light data. The Landsat 8 data, spanning the year 2019, were aggregated into a median composite to mitigate the influence of atmospheric conditions and seasonal variations. The decision to use a one-year median, akin to the established three-year median employed in contemporary state-of-the-art poverty prediction models, such as the one by Pettersson et al. \cite{pettersson2023time}, reflects a desire to explore within-year fluctuations, making it a more suitable benchmark for comparative analysis. This approach also acknowledges the potential loss of significant temporal insights that may occur when utilizing broader averages. To evaluate the proposed method, we combined the Landsat 8 data with intra-annual NDVI values or wavelet-extracted features.

\subsection*{Model and resources}

A dataset containing 42,053 images from 36 African countries was compiled. To reduce the impact of seasonal variations, Landsat-based images were generated using the per-band annual median values from all cloud-free pixels, denoted \( \bar{p}[g_{i}, s_{Landsat}] \). Each image is sized at \(224 \times 224\) pixels, where each pixel corresponds to a \(30 \times 30\) meter area on the Earth's surface, covering a total area of \(6.72 \text{ km} \times 6.72 \text{ km}\). The dataset was randomly split into training, validation, and test sets with proportions of 60\%, 20\%, and 20\%, respectively. These datasets were downloaded from GEE \cite{amani2020google} and transferred to the National Academic Infrastructure for Supercomputing in Sweden (NAISS).

The VGG-16 model served as the deep learning model. To ensure compatibility with our problem, an additional convolutional layer was incorporated into the model’s architecture, allowing transformation of the multi-dimensional data into the required single scalar value. Different VGG-16 models were trained for 100 epochs, each using distinct subsets of the dataset as follows:

\begin{itemize}
    \item Base model using 7 raw Landsat images \( \bar{p}[g_{i}, s_{Landsat}] \): This model serves as the baseline and comprises seven bands derived from the annual median Landsat data. The approach is based on existing literature that employs yearly or multi-year aggregations for data preparation in wealth analysis, which is why it is referred to as the base model.
    \item Base model with temporal NDVI, which includes 7 raw Landsat images \( \bar{p}[g_{i}, s_{Landsat}] \) and 12 raw NDVI images \( q[g_{i}, \tau, s_{NDVI}] \): Building on the findings from previous literature, we examine whether adding the twelve monthly NDVI averages as additional images to the base model provides meaningful insights in our application.
    \item Full dataset \(q[g_{i}, \tau, s_{Landsat}]\) includes monthly Landsat data with 7 bands per month, resulting in \(12 \times 7 = 84\) bands. This approach does not align with practical requirements, as much of the literature utilizes yearly or three-year median data to create mosaic images that cover the study area with minimal spatial gaps. While generating a yearly median dataset is challenging, using monthly data for wealth studies is generally considered impractical. However, we aim to analyze whether having monthly data, similar to the base model but with monthly availability, provides additional informative value for our application.
    \item  Statistical wavelet-based features from intra-annual NDVI images, including three from low-frequency compositions \( LFC_{stats}{[q[g_{i}, \tau, s_{NDVI}]]} \) and three from high-frequency compositions \( HFC_{stats}{[q[g_{i}, \tau, s_{NDVI}]]} \): The 6 statistical wavelet-based feature images are generated by summarizing the coefficients resulting from the wavelet transformation applied to the 12 NDVI images. This process involves extracting the minimum, maximum, and mean values of high- and low-frequency coefficients separately. In the dataset, \( LFC_{stats} \) represents the statistical summaries (minimum, maximum, and mean values) of the low-frequency coefficients, while \( HFC_{stats} \) represents the statistical summaries of the high-frequency coefficients.
    \item Base model with 7 raw Landsat images \( \bar{p}[g_{i}, s_{Landsat}] \) and 6 statistical wavelet-based features from intra-annual NDVI images, including three from low-frequency components \( LFC_{stats}{[q[g_{i}, \tau, s_{NDVI}]]} \) and three from high-frequency components \( HFC_{stats}{[q[g_{i}, \tau, s_{NDVI}]]} \): The use of statistical features aims to capture information that is not time-dependent, unlike the original NDVI data, which may be influenced by seasonality and thus may not be sufficiently informative. In contrast, statistical features derived from wavelet transformations could provide more robust and meaningful insights for the analysis.
    \item Base model with 7 raw Landsat images \( \bar{p}[g_{i}, s_{Landsat}] \) and one of six wavelet-based NDVI features (either from low-frequency \( LFC_{stats}{[q[g_{i}, \tau, s]]} \) or high-frequency components \( HFC_{stats}{[q[g_{i}, \tau, s]]} \)): This approach aims to investigate whether the each statistical summary feature is informative for the model while exploring the potential to reduce the number of additional images to fewer than six.
\end{itemize}

\subsection*{Evaluation metrics}

The performance of the trained deep models are assessed using the R-squared ($R^2$) score, known as coefficient of determination. The $R^2$ score is a statistical measure that quantifies the proportion of the variance in the dependent variable that is predictable from the independent variables in a regression model. Specifically, it measures the goodness of fit of the model to the observed data. The $R^2$ score can be described in equation \ref{eq:R2}.

\begin{equation}
    R^2 = 1- \frac{RSS}{TSS}
    \label{eq:R2}
\end{equation}

where Residual sum of squares $RSS$, and Total sum of squares $TSS$ are defined in equations \ref{eq:RSS} and \ref{eq:TSS}, respectively. 

\begin{equation}
    RSS = \sum_{i=1}^n(y_i-\hat{y}_i)^2
    \label{eq:RSS}
\end{equation}

\begin{equation}
    TSS = \sum_{i=1}^n(y_i-\bar{y})^2
    \label{eq:TSS}
\end{equation}

where $y_i$ is the corresponding true value of the $i$-th sample, $\hat{y}_i$ is the predicted value and $\bar{y}$ is the mean value of the observed data. A high $R^2$ score, close to 1, indicates that the model explains a large portion of the variability in the data, suggesting a good fit. Evaluating our model's performance using the $R^2$ score will provide valuable insights into its predictive accuracy and effectiveness in capturing the underlying relationships within the data.

The $R^2$ scores for each model are presented in Table \ref{tab:r2_scores}. The base model, trained solely on the seven bands from the annual median Landsat data, achieved an $R^2$ score of 0.69, serving as the baseline. In comparison, the model utilizing the full dataset of 12 images per each of the 7 Landsat bands performed the best, with an $R^2$ score of 0.85, highlighting the effectiveness of incorporating comprehensive temporal information. The base model augmented with temporal monthly NDVI values with $R^2$ score of 0.75 demonstrates the potential of using temporal data for more accurate poverty analysis. However, from a practical perspective, temporal NDVI data is not limited to Landsat and can also be sourced from datasets like MODIS. Additionally, the model incorporating additional raw NDVI images serves as a comparison to methods based on statistical summaries of NDVI data.

The statistical wavelet-based features from temporal NDVI images were used to train a model for poverty analysis, achieving an $R^2$ score of 0.66. This score is lower than that of the base model, indicating that the spectral bands are currently more critical for model performance than temporal resolution. Nonetheless, comparing the full dataset with the base model highlights that there is potential for improvement by incorporating intra-annual temporal variations to enhance poverty estimation accuracy.

The additional image bands proved effective in summarizing and enhancing the performance of the models. The wavelet-based model, utilizing six temporal summary images, demonstrated notable improvements over the baseline, achieving an $R^2$ score of 0.79. This result underscores the efficacy of the wavelet-based time-series summary approach in capturing valuable temporal dynamics. Consequently, features robust to seasonality, like those derived from wavelet-based statistical summaries, provide more informative input for training deep learning models.

Moreover, models that incorporated even a single additional image band, specifically using HFC features, showed substantial enhancements compared to the base model. Notably, models employing maximal and average high-frequency coefficients achieved \(R^2\) scores of 0.77 and 0.79, respectively. However, using low-frequency features did not impact accuracy. This is understandable, as low-frequency coefficients are less informative in this context; they are more similar to the yearly median dataset \( \bar{p}[g_{i}, s_{Landsat}] \) and can be considered redundant features.

\begin{table*}[t]
\centering
\caption{Test $R^2$ score for the simulated dataset, with the target variable consistently set as the average nighttime light data across all setups. The "Dataset description" and "Dataset" columns indicate the data used to train the VGG-16 models.}
\label{tab:r2_scores}
\begin{tabular}{|l|l|l|} \hline
{\textbf{Dataset description}}  & \textbf{Dataset} & \textbf{$R^2$ score}  \\ \hline

Base model using 7 raw Landsat images & \( \bar{p}[g_{i}, s_{Landsat}] \) & 0.69  \\ 
Base model and temporal NDVI & \( \bar{p}[g_{i}, s_{Landsat}] + q[g_{i},\tau, s_{NDVI}] \) & \textbf{0.75}  \\ 
Full dataset with raw Landsat images per month & \(q[g_{i},\tau, s_{Landsat}]\) & 0.85 \\ \hline

Statistical wavelet-based features from temporal NDVI images & \( LFC_{stats} + HFC_{stats} \) & 0.66 \\
Base model and 6 statistical wavelet-based features & \( \bar{p}[g_{i}, s_{Landsat}] + LFC_{stats} + HFC_{stats} \) & \textbf{0.79} \\
Base model and max values of HFC & \( \bar{p}[g_{i}, s_{Landsat}] + HFC_{stats=max} \) & \textbf{0.77} \\ 
Base model and mean values of HFC & \( \bar{p}[g_{i}, s_{Landsat}] + HFC_{stats=min} \) & 0.72 \\ 
Base model and min values of HFC & \( \bar{p}[g_{i}, s_{Landsat}] + HFC_{stats=mean} \) & \textbf{0.79} \\ 
Base model and max values of LFC & \( \bar{p}[g_{i}, s_{Landsat}] + LFC_{stats=max} \) & 0.68 \\ 
Base model and mean values of LFC & \( \bar{p}[g_{i}, s_{Landsat}] + LFC_{stats=min} \) & 0.67 \\ 
Base model and min values of LFC & \( \bar{p}[g_{i}, s_{Landsat}] + LFC_{stats=mean} \) &  0.67 \\ \hline

\hline

\hline
\end{tabular}
\end{table*}

\subsection*{DHS simulation}

Given the strong performance of our summarization method in a simulated setting using the mean value from nighttime light data as the output, we tested our approach in an applied setting using DHS\cite{DHS_Program} data. To address the issue of missing data, we used a 3-year median for each month instead of a 1-year median as our input data. Additionally, nighttime light imagery was included as part of our input. The International Wealth Index (IWI) served as the output measure of poverty. We utilized only complete data points with no missing pixels, resulting in a total of 8598 samples. The remaining setup remained unchanged, and the 5-fold cross-validation test scores are presented in Table \ref{tab:r2_scores_dhs}.

\begin{table*}[t]
\centering
\caption{Test $R^2$ score for DHS datasets, with the target variable being the IWI from the DHS dataset across all setups. The "Dataset description" and "Dataset" columns indicate the data used to train the VGG-16 models.}
\label{tab:r2_scores_dhs}
\begin{tabular}{|l|l|l|} \hline
{\textbf{Dataset description}}  & \textbf{Dataset} & \textbf{$R^2$ score}  \\ \hline

Base model using 7 raw Landsat images and NL data & \( \bar{p}[g_{i}, s_{Landsat}] + \bar{p}[g_{i}, s_{NL}] \) & 0.81  \\ 
Base model with temporal NDV &  \( \bar{p}[g_{i}, s_{Landsat}] + \bar{p}[g_{i}, s_{NL}] + q[g_{i},\tau, s_{NDVI}]\) & 0.83  \\ \hline

Base model and 6 statistical wavelet-based features &  \( \bar{p}[g_{i}, s_{Landsat}] + \bar{p}[g_{i}, s_{NL}] + LFC_{stats} + HFC_{stats}\) & 0.83 \\
Base model and max values of HFC &  \( \bar{p}[g_{i}, s_{Landsat}] + \bar{p}[g_{i}, s_{NL}] + HFC_{stats=max} \) &  0.81 \\ 
Base model and mean values of HFC & \( \bar{p}[g_{i}, s_{Landsat}] + \bar{p}[g_{i}, s_{NL}] + HFC_{stats=mean} \) &  0.82 \\ 
Base model and min values of HFC &  \( \bar{p}[g_{i}, s_{Landsat}] + \bar{p}[g_{i}, s_{NL}] + HFC_{stats=min} \) &  0.82 \\ 
Base model and max values of LFC  &  \( \bar{p}[g_{i}, s_{Landsat}] + \bar{p}[g_{i}, s_{NL}] + LFC_{stats=max}\) &  \textbf{0.84} \\ 
Base model and mean values of LFC & \( \bar{p}[g_{i}, s_{Landsat}] + \bar{p}[g_{i}, s_{NL}] + LFC_{stats=mean}\) &  0.82 \\ 
Base model and min values of LFC &  \( \bar{p}[g_{i}, s_{Landsat}] + \bar{p}[g_{i}, s_{NL}] + LFC_{stats=min}\) &  0.81 \\ \hline

\hline
\end{tabular}
\end{table*}

We observed that using the base model along with the maximum of LFC improved the model's performance, achieving an \(R^2\) score of 0.84, which outperformed all other models. However, since the evaluation is based on the perturbed DHS dataset and the accuracy ranges vary only in the second decimal place, we should not overinterpret these results. This finding simply reinforces the value of utilizing temporal NDVI features in socioeconomic studies, as supported by simulation data. Finally, since adding these features to the base model does not adversely affect accuracy, they should be considered in future studies conducted in more controlled environments.

\section*{Discussion}

The results of this study demonstrate the significant advantage of incorporating temporal information into models for poverty estimation using satellite imagery. While the baseline model, using only the seven bands from yearly median Landsat data, achieved a test $R^2$ score of 0.69, models that included additional temporal images showed substantial improvements, with the best model reaching an $R^2$ score of 0.85 for the full dataset in simulated setting.

More importantly however, this research highlights the effectiveness of wavelet-based summarization in capturing those temporal dynamics. Model using six constructed temporal images achieved $R^2$ scores of 0.79. In applied setting utilizing DHS dataset, while less significant, a benefit in $R^2$ scores was also repeated. This indicates that wavelet transforms can distill essential temporal information effectively, achieving performance close to that of models utilizing full temporal datasets and performs better than model incorporating NDVI images.

The ability of wavelet summarization to maintain high model performance while significantly reducing the amount of additional data required is particularly noteworthy. This approach allows for substantial computational resource savings in terms of storage and training. Even with a single additional image derived from wavelet high frequency coefficients, models showed significant improvements over the baseline, underscoring the efficiency and utility of wavelet-based summarization of intra-annual NDVI data.

In summary, this research highlights the potential of wavelet-based methods to optimize an effective balance between performance and resource efficiency, making them a valuable tool for remote sensing and related applications. In this study, we focused on NDVI from Landsat data, but this approach could be extended to include other sources like MODIS. Combining spatial and frequency domain information from Landsat with intra-annual data from MODIS NDVI presents an exciting opportunity for future research. Moving forward, it will be important to further explore and refine these techniques across various model architectures to maximize their utility and efficiency.

\section*{Methods}

The central aim of this work is to explore the feasibility and effectiveness of utilizing wavelet-based intra-annual feature extraction for training a deep learning architecture in poverty estimation. The proposed method posits that intra-annual temporal variation is informative for poverty estimation and wealth index analysis. By employing a wavelet-based feature extraction method, intra-annual variation features are incorporated into machine learning models. The experimental approach involves transforming time-series satellite images into their wavelet representations before integrating them into the deep learning model in a simulated setting. This work employs the VGG-16 \cite{vgg16} deep learning architecture to investigate multiple approaches for summarizing the data.

Despite the availability of other potential summarization techniques, NDVI has been extensively used in previous poverty analysis studies, demonstrating its effectiveness in capturing socio-economic dynamics and environmental conditions. While NDVI doesn't explicitly consider socio-economic factors, it often becomes part of models predicting these variables. For instance, Sedda et al. \cite{sedda2015} utilized spatial statistics techniques to demonstrate an inverse relationship between NDVI and poverty intensity in West Africa. Moreover, NDVI is positively correlated with improved child survival rates, nutrition, and anthropometric indicators like wasting, owing to its positive influence on crop yields \cite{JOHNSON2014}. Similarly, Tang et al. \cite{tang2018dynamic} demonstrated that publicly available, moderate-resolution vegetation indices can be effectively utilized with convolutional neural networks to generate precise poverty estimates for developing countries reliant on agriculture. This is further supported by work of Maciel et al.'s \cite{land_use}, that employed the Enhanced Vegetation Index (EVI), akin to NDVI, for extracting time series satellite data to discern land change events like deforestation or agricultural expansion.

By leveraging NDVI as our primary input data, we align with existing literature emphasizing its significance in socioeconomic analyses. Notably in this work, Tang, Liu, and Matteson's \cite{tang2022predicting} work on predicting poverty with vegetation index stands out. Leveraging a VGG-16 model, they mapped NDVI images to corresponding nighttime light intensities, thereby establishing a crucial link between vegetation health and socioeconomic indicators. While NDVI primarily focuses on vegetation health and density, it still captures broader landscape changes that are relevant to socioeconomic factors, such as changes in land cover and land use patterns. By employing this approach, they extracted comprehensive features from NDVI imagery and adeptly employed regression models to forecast key poverty metrics, including the logarithm of consumption expenditure per capita and wealth index. This influenced our decision to employ NDVI for capturing intra-annual variation in poverty prediction.

In this paper, we denote a geographical location on Earth as  \( g \). Each pixel within the satellite data is denoted as \( p \), and it corresponds to a specific location \( p(g_{i}) \), where \( g_{i} \) represents one of the discretized geographical points. The discretization of geographical locations stems from the fundamental nature of satellite imagery, which presents data in the form of images. In this context, each pixel within the satellite imagery serves as a representation of a specific portion of the Earth's surface. In this case, each discretized geographical point \( g_{i} \) delineates a square area of 30 by 30 meters on the Earth's surface.

Moreover, it is considered that each location \( g_{i} \) is visited \( T_{g_{i},s} \) times per year by a specific satellite $s$. Thus, the pixel value is treated as a signal dependent on location, time, and satellite imagery program. Specifically, \( p(g_{i},t,s) \) represents the pixel value at location \( g_{i} \) at time \( t \) captured by satellite $s$. To focus on annual patterns, this analysis is restricted to satellite data spanning one year, allowing to extract intra-annual features.

Assuming the pixel value at a desired location $i$, captured by a specific satellite $s$ is influenced by temporal factors, such as weather conditions, which may introduce non-robust features. To mitigate these fluctuations, monthly averaging is employed as a smoothing technique. This approach aligns with studies that have utilized monthly MODIS NDVI datasets \cite{karthikeyan2013predictability,martinez2009vegetation,yan2022comparison}. The computation of the average signal over a one-month interval can be represented by using the notation in equation \ref{eq:Q_month}.

\begin{equation}
    q[g_{i},\tau, s] = \{\frac{1}{T_{g_{i}{,s,j}}} 
    \sum_{t=1}^{T_{g_{i}{,s,j}}} p_j(g_{i},s,t) , j = 1, \ldots, 12\}
    \label{eq:Q_month}
\end{equation}

Where:
\begin{itemize}
    \item \( q[g_{i},\tau,s] \) denotes the average signal value of the pixel over a monthly interval of the specific satellite. While both \( p \) and \( q \) consist of discrete values, the primary focus is on \( q \) as the discrete signal due to its fixed length of 12, unlike \( p \) which lacks a consistent length. This is why we used the notation \( q \) that shows the signal is further quantized to 12 values.
    \item The index \( j \) ranges from 1 to 12, representing each monthly interval within a consecutive one-year period.
    \item \(T_{g_{i}{,s,j}}\) signifies the number of samples within a monthly interval.
    \item \(p_j(g_{i},s,t) \) represents the pixel value of satellite $s$ in month \( j \) at time \( t \) at location $g_i$.
\end{itemize}

On the other hand, most literature \cite{yeh2020using,pettersson2023time} used at least yearly, and sometime to 3-year median, to achieve a data inputting to a ML model to predict the poverty level. We show this data as $\bar{p}[g_{i}, s]$ which shows the median on data at location $g_i$ captured by satellite $s$.

\begin{equation}
    \bar{p}[g_{i}, s] = \{\frac{1}{T_{g_{i}{,s}}} 
    \sum_{t=1}^{T_{g_{i}{,s}}} p(g_{i},s,t) 
\end{equation}

Wavelets serve as a powerful tool in time-series analysis in signal processing. A wavelet is a wavelike oscillation that is concentrated within a specific time interval, and it can represent signal domain-frequency information for analysis. They can be used to distill important features from complex datasets, capturing both broad trends and fine details. For digital signal processing, such as time-series satellite data, the Discrete Wavelet Transform (DWT) is defined as:

\begin{equation}
D[a,b]=\frac{1}{\sqrt{b}}\sum_{m=0}^{p-1}f[t_{m}]\psi \left [ \frac{t_{m}-a}{b} \right]
\end{equation}

where:
\begin{itemize}
    \item $D[a,b]$ represents the wavelet-transformed signal in discrete form.
    \item $f[t_{m}]$ is the discrete input signal at sample points $t_m$. Different wavelet filters, such as Daubechies, Morlet, and Haar, are employed based on their ability to provide sparse representations and capture specific signal characteristics.
    \item The parameters $a$ and $b$ are the discrete translation and scale parameters, respectively, where $a, b \in \mathbb{Z}$. For practical applications, these parameters are typically discretized as $a=k2^{-j}$ and $b=2^{-j}$, where $j, k \in \mathbb{Z}$ are referring to scale index and wavelet-transformed signal index, respectively \cite{srivastava2018wavelet}.
\end{itemize}

The wavelet transform of \( q[g_{i},s,\tau] \) is computed through convolution with a Meyer wavelet function. The selection of the Meyer wavelet, distinguished by its larger filter length of 62, was motivated by its demonstrated efficacy in time series analysis of vegetation dynamics derived from NDVI \cite{MARTINEZ20091823}. 

In this specific study case, the datasets consist of:
\begin{itemize}
\item \( \bar{p}[g_{i}, s] \): Seven bands derived from yearly median Landsat data. These bands are independent of time, as the median value is computed for each pixel.
\item \( q[g_{i},\tau,s] \): Twelve monthly averages of NDVI values coming from Landsat or MODIS satellite imagery program. In this work NDVI is derived from Landsat data which determines $s$, \( g_{i} \) represents the pixel, and \( \tau \) represents the month. 
\item Low-frequency (approximation) coefficients from the wavelet transformation are denoted as \( LFC \):

\begin{equation}
    LFC[q[g_{i},\tau,s]] = \sum_{\tau=1}^{12} q[g_{i},\tau,s] LP[2n-\tau]
    \label{eq:LFC}
\end{equation}

\item High-frequency (detail) coefficients from the wavelet transformation are denoted as \( HFC \):

\begin{equation}
    HFC[q[g_{i},\tau,s]] = \sum_{\tau=1}^{12} q[g_{i},\tau,s] HP[2n-\tau]
    \label{eq:HFC}
\end{equation}

\end{itemize}

However, wavelet transformation alone does not yield a more concise representation than the original 12 images. Utilizing summary statistics derived from wavelet-transformed satellite signal can potentially enhance model effectiveness with fraction of the initial information. Studies across diverse domains illustrate this. For example, Phinyomark et al. \cite{Phinyomark_Nuidod_Phukpattaranont_Limsakul_2012} improved class separability and classification accuracy in surface electromyography analysis by employing statistical measures like mean absolute value and zero crossing. Shah et al. \cite{app14020599} achieved superior seizure detection in EEG signals by incorporating features such as mean value and standard deviation from the wavelet transformation. Rostaminia et al. \cite{10.1145/3513023} demonstrated the efficacy of summary statistics in sleep monitoring, outperforming commercial wearables with features like maximum, mean, and standard deviation.

\begin{figure}[H]
    \centering
    \includegraphics[width=0.95\linewidth]{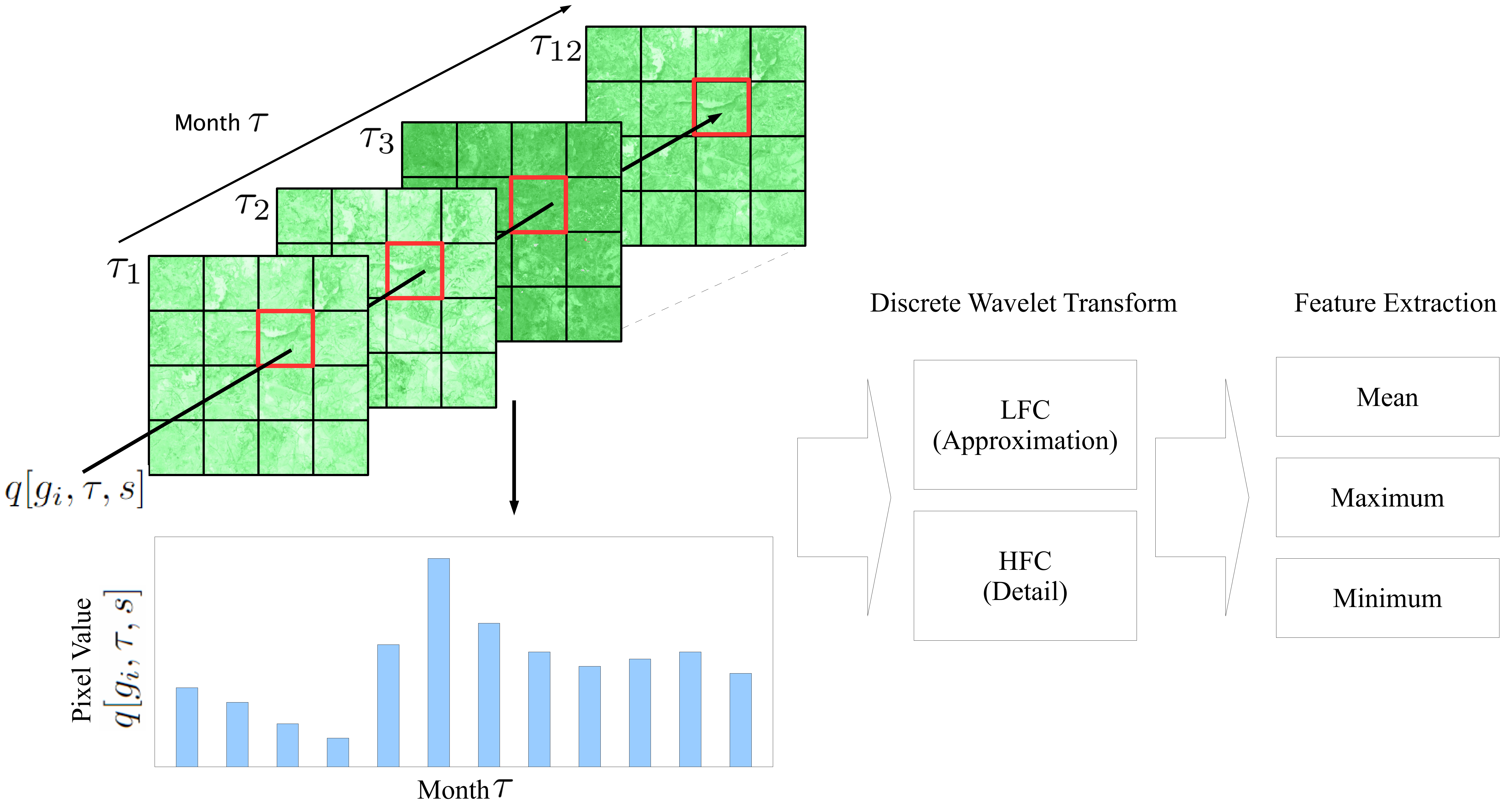}
    \caption{Data pre-processing for DWT-based feature extraction}
    \label{fig:method_shema}
\end{figure}

In our setup similar strategy is applied. An additional set of six images was generated from the twelve monthly averages of NDVI, aiming to investigate the potential for synthesizing temporal data within this dataset. Proposed pre-processing procedure is illustrated in figure \ref{fig:method_shema}. 

\begin{figure}[H]
    \centering
    \includegraphics[width=1\linewidth]{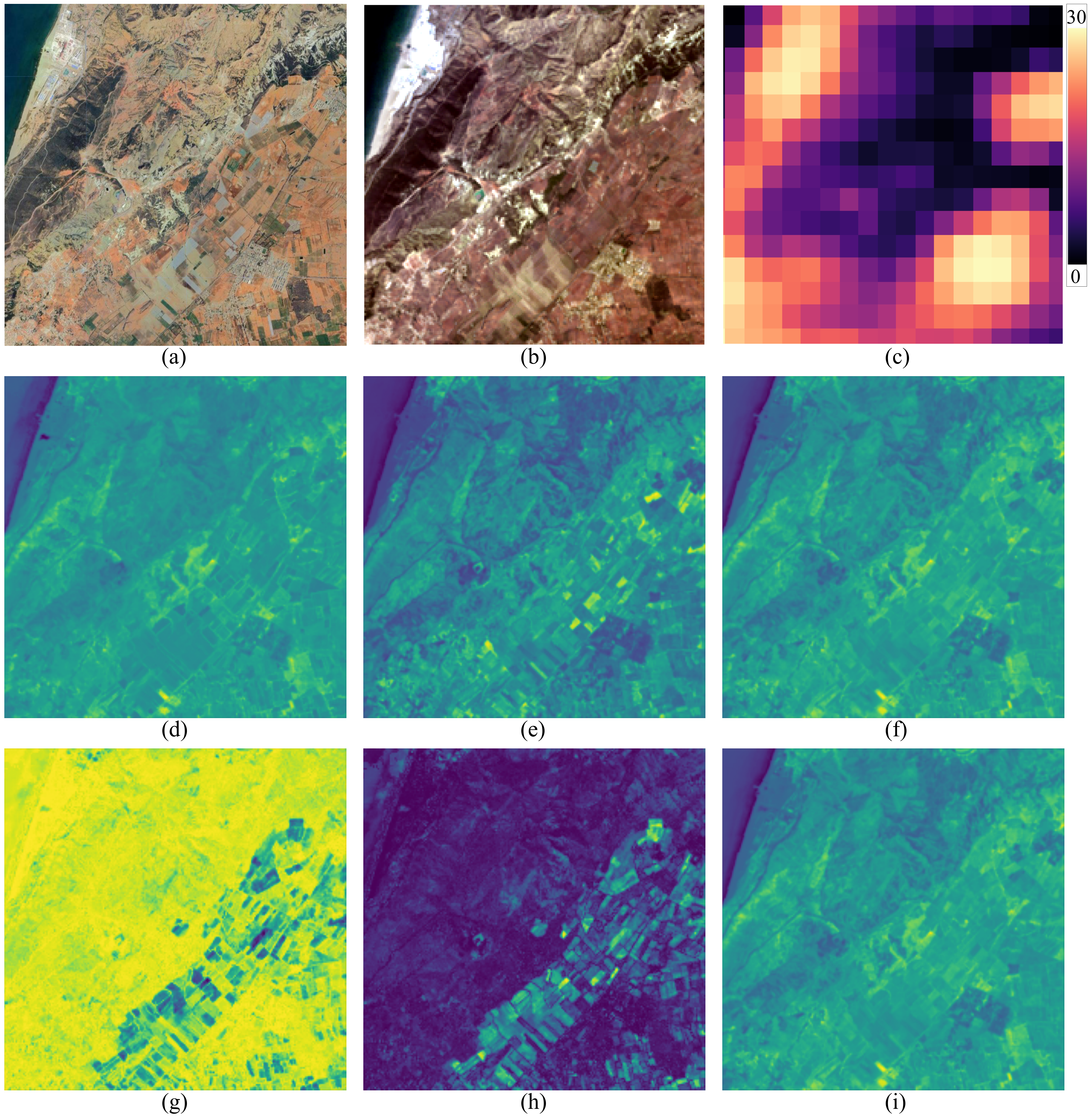}
    \caption{Example tile centered at Longitude 0.1515 and Latitude 35.996 in Algeria. (a) High-resolution Google imagery providing detailed visualization of the area. (b) Landsat-8 true color (RGB) visualization. (c) VIIRS nighttime light data. (d) Minimum LFC value. (e) Maximum LFC value. (f) Mean LFC value. (g) Minimum HFC value. (h) Maximum HFC value. (i) Mean HFC value.}
    \label{fig:waveletfeatures}
\end{figure}

For each corresponding pixel \( q[g_{i},\tau,s] \), a temporal signal comprising 12 samples was extracted, characterizing the changes occurring at specific locations over time. A single-level 1D DWT decomposition was subsequently conducted to derive both LFC and HFC from each signal using equations \ref{eq:LFC} and \ref{eq:HFC}. Summary statistics, encompassing the mean, minimal, and maximal values of these coefficients, were computed. The resulting values were then utilized directly as pixel intensities for the newly generated images. To illustrate using an example, for tile shown in figure \ref{fig:waveletfeatures}, six summary images have been generated: three for LFC and three for HFC.

Figure \ref{fig:waveletfeatures}(a) presents a high-resolution view of the study area. In Figure \ref{fig:waveletfeatures}(b), a lower-resolution Landsat-8 RGB visualization shows a similar texture. Figure \ref{fig:waveletfeatures}(c) displays VIIRS nighttime light data, which aids in identifying human settlement areas through visual inspection. The wavelet-based intra-annual NDVI features in Figures \ref{fig:waveletfeatures}(d) and \ref{fig:waveletfeatures}(e) emphasize nearby cropland areas. In particular, the Maximum of LFC, Minimum of HFC, and Maximum of HFC highlight these cropland areas, which are valuable features for downstream socioeconomic analysis. This demonstrates the potential of these features to provide meaningful insights for this study.

These features, together with other signals like \(\bar{p}[g_{i}, s]\), are used to assess whether these intra-annual wavelet-based features provide valuable insights for economic studies and poverty estimation.

\section*{Data availability}
The Earth observation data, including the Landsat and VIIRS nighttime light datasets, were obtained through the Google Earth Engine platform. The links below provide access to these data collections:

Landsat dataset: \url{https://developers.google.com/earth-engine/datasets/catalog/LANDSAT_LC08_C02_T1_L2}

VIIRS Nighttime Light dataset: \url{https://developers.google.com/earth-engine/datasets/catalog/NOAA_VIIRS_DNB_MONTHLY_V1_VCMCFG}

The Demographic and Health Survey (DHS) dataset is accessible via the following link:

DHS dataset: \url{https://dhsprogram.com/data/available-datasets.cfm}

Both data sources are publicly accessible but require user registration.

For the simulated dataset, points were generated in human settlement areas across the African continent. These points were used to link the Landsat-8 data to the average nighttime light data in those areas. The generated points are available in the GitHub repository, which can be found in the code availability section.

\section*{Code availability}
The results were produced using Python (v3.10.12) on the National Academic Infrastructure for Supercomputing in Sweden (NAISS) and Google Earth Engine (GEE). More information about the scripts are available on GitHub in the repository of the AI and Global Development Lab: \hyperlink{https://github.com/AIandGlobalDevelopmentLab/IntraAnnual-NDVI-wavelet-Poverty}{https://github.com/AIandGlobalDevelopmentLab/IntraAnnual-NDVI-wavelet-Poverty}.

\bibliography{refs}


\section*{Acknowledgements (not compulsory)}

This work was supported by the Swedish Research Council
through The AI and Global Development Laboratory under Grant 2020-03088 and Grant 2020-00491.

The computations and data handling were enabled by resources provided by the National Academic Infrastructure for Supercomputing in Sweden (NAISS), partially funded by the Swedish Research Council through grant agreement no. 2022-06725.

\section*{Author contributions statement}

Conceptualization, M.K. and A.D.; Methodology, M.K., and K.S.; Investigation, M.K., K.S., and A.D.; Writing—original draft, M.K., and K.S.; Writing—review and editing, M.K., K.S. and A.D.; Supervision, M.K., and A.D.; Funding acquisition, A.D. All authors have read and agreed to the published version of the manuscript.




\end{document}